\documentclass[sigconf]{acmart}
\AtBeginDocument{%
  }

\copyrightyear{2026}
\acmYear{2026}
\setcopyright{cc}
\setcctype{by}
\acmConference[IUI Companion '26]{Companion Proceedings of the 31st International Conference on Intelligent User Interfaces}{March 23--26, 2026}{Paphos, Cyprus}
\acmBooktitle{Companion Proceedings of the 31st International Conference on Intelligent User Interfaces (IUI Companion '26), March 23--26, 2026, Paphos, Cyprus}
\acmPrice{}
\acmDOI{10.1145/3742414.3794777}
\acmISBN{979-8-4007-1985-1/2026/03}

\usepackage{enumerate}

\begin{document}

\title{CausalAgent: A Conversational Multi-Agent System for End-to-End Causal Inference}

\author{Jiawei Zhu}
\email{3123005571@mails.gdut.edu.cn}
\authornote{Both authors contributed equally to this research.}
\orcid{0009-0006-6516-3611}
\affiliation{%
  \institution{Guangdong University of Technology}
  \city{Guangzhou}
  \state{Guangdong}
  \country{China}
}

\author{Wei Chen}
\authornotemark[1]
\email{chenweidelight@gmail.com}
\authornote{Corresponding author.}
\affiliation{%
  \institution{Guangdong University of Technology}
  \city{Guangzhou}
  \state{Guangdong}
  \country{China}
}

\author{Ruichu Cai}
\email{cairuichu@gmail.com}
\affiliation{%
  \institution{Guangdong University of Technology}
  \city{Guangzhou}
  \state{Guangdong}
  \country{China}
}
\affiliation{%
  \institution{Peng Cheng Laboratory}
  \city{Shenzhen}
  \country{China}
}

\renewcommand{\shortauthors}{Zhu, Chen and Cai, et al.}

\begin{abstract}
Causal inference holds immense value in fields such as healthcare, economics, and social sciences. However, traditional causal analysis workflows impose significant technical barriers, requiring researchers to possess dual backgrounds in statistics and computer science, while manually selecting algorithms, handling data quality issues, and interpreting complex results. To address these challenges, we propose CausalAgent, a conversational multi-agent system for end-to-end causal inference. The system innovatively integrates Multi-Agent Systems (MAS), Retrieval-Augmented Generation (RAG), and the Model Context Protocol (MCP) to achieve automation from data cleaning and causal structure learning to bias correction and report generation through natural language interaction. Users need only upload a dataset and pose questions in natural language to receive a rigorous, interactive analysis report. As a novel user-centered human-AI collaboration paradigm, CausalAgent explicitly models the analysis workflow. By leveraging interactive visualizations, it significantly lowers the barrier to entry for causal analysis while ensuring the rigor and interpretability of the process.
\end{abstract}

\begin{teaserfigure}
  \includegraphics[width=\textwidth]{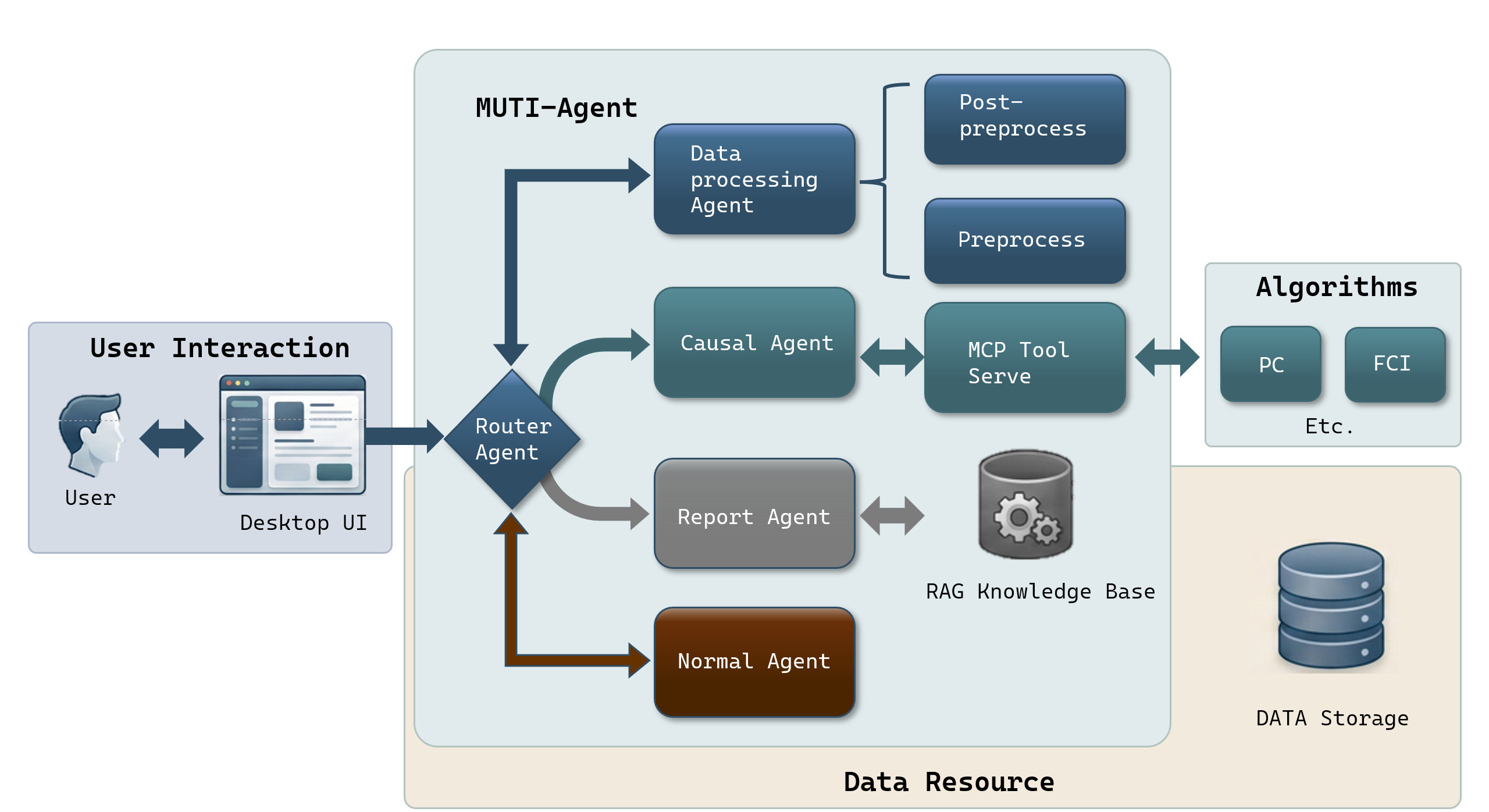}
  \caption{Overview of the CausalAgent Interface and Workflow.}
  \Description{A screenshot showing the chat interface on the left and the interactive causal graph on the right.}
  \label{fig:fig1}
\end{teaserfigure}

\maketitle

\section{Introduction}
Causal inference is crucial for decision-making in many fields like healthcare \cite{pearl2009causality}, economics \cite{imbens2015causal}, Neuroscience~\cite{cai2024granger}, and financial analysis \cite{chen2025higher}. However, traditional workflows impose high barriers, demanding researchers master complex algorithms and manually address engineering challenges \cite{spirtes2000causality, chen2024identification}. While Large Language Models (LLMs) offer potential for causal reasoning, single models struggle with hallucination in long-process tasks \cite{ji2023survey,chen2025causal}. Moreover,existing tools (e.g., DoWhy \cite{sharma2020dowhy}, Causal-learn \cite{zheng2024causal}) often lack intuitive, end-to-end automation.

To address these challenges, we introduce \textbf{CausalAgent}\footnote{Source code: \url{https://github.com/DMIRLAB-Group/CausalAgent}}, a system encapsulating the causal analysis process into collaborative agent services \cite{xi2023rise}. Users interact via natural language to complete a closed loop: from data processing and RAG-based knowledge retrieval \cite{lewis2020rag} to causal structure learning and report generation. By explicitly modeling the workflow and employing Supervised Fine-Tuning (SFT), CausalAgent mitigates hallucination risks and ensures rigorous, reproducible analysis.

In this demonstration, we showcase how CausalAgent guides users through the entire causal discovery pipeline—from raw data processing to interactive, multi-round causal reasoning—via a unified conversational interface.

\section{System Overview}
\subsection{Architecture}
The core of CausalAgent lies in its highly scalable Multi-Agent System (MAS). As illustrated in Figure \ref{fig:fig1}, the system follows a modular design utilizing LangGraph \cite{langchain2024langgraph} for intelligent routing. The architecture is composed of three primary components: \textbf{Data Processing Agent}, \textbf{Causal Structure Learning Agent}, and \textbf{Reporting Agent}. Each agent focuses solely on its own logic and I/O constraints, preserving context information within the shared state to ensure consistency across different analysis stages \cite{wu2023autogen}. The specific details of these components are provided in the following subsections.

\subsection{Data Processing Agent}

This agent handles both preprocessing and quality detection.
\textbf{Preprocessing:} It validates uploaded files, summarizes statistics (e.g., missing rates, unique values), and infers the meaning from variables' names. The system provides a ``causal analysis friendliness" rating and generates visualizations (histograms, correlation heatmaps) to highlight data quality issues.
\textbf{Quality Detection:} It utilizes graph algorithms to detect cycles violating the Directed Acyclic Graph (DAG) assumption \cite{pearl2009causality}. If cycles are found, the agent proposes edge modifications based on domain priors and statistical associations, requiring the LLM to provide confidence-weighted decisions.
\subsection{Causal Structure Learning Agent}

This agent acts as an algorithm scheduler based on natural language prompts in the state and knowledge priors, utilizing the Model Context Protocol (MCP) \cite{anthropic2024mcp}. It invokes causal learning algorithms from the tool server to construct causal graphs. Currently, the algorithm library integrates the PC algorithm\footnote{Source code: \url{https://github.com/py-why/causal-learn}} \cite{spirtes2000causality} for searching DAG structures based on conditional independence tests, and OLC-based algorithms\footnote{Source code: \url{https://github.com/DMIRLAB-Group/CDMIR}} \cite{cai2023causal,chen2024identification} for handling latent confounders. The agent selects the most appropriate algorithm based on natural language instructions and data characteristics, outputting a node list and adjacency matrix. The causal graph structure generated in this phase serves as the core input for post-processing and report generation modules, and is persisted as a checkpoint to facilitate subsequent multi-turn inquiries.

\begin{figure*}
  \centering
  \includegraphics[width=\textwidth]{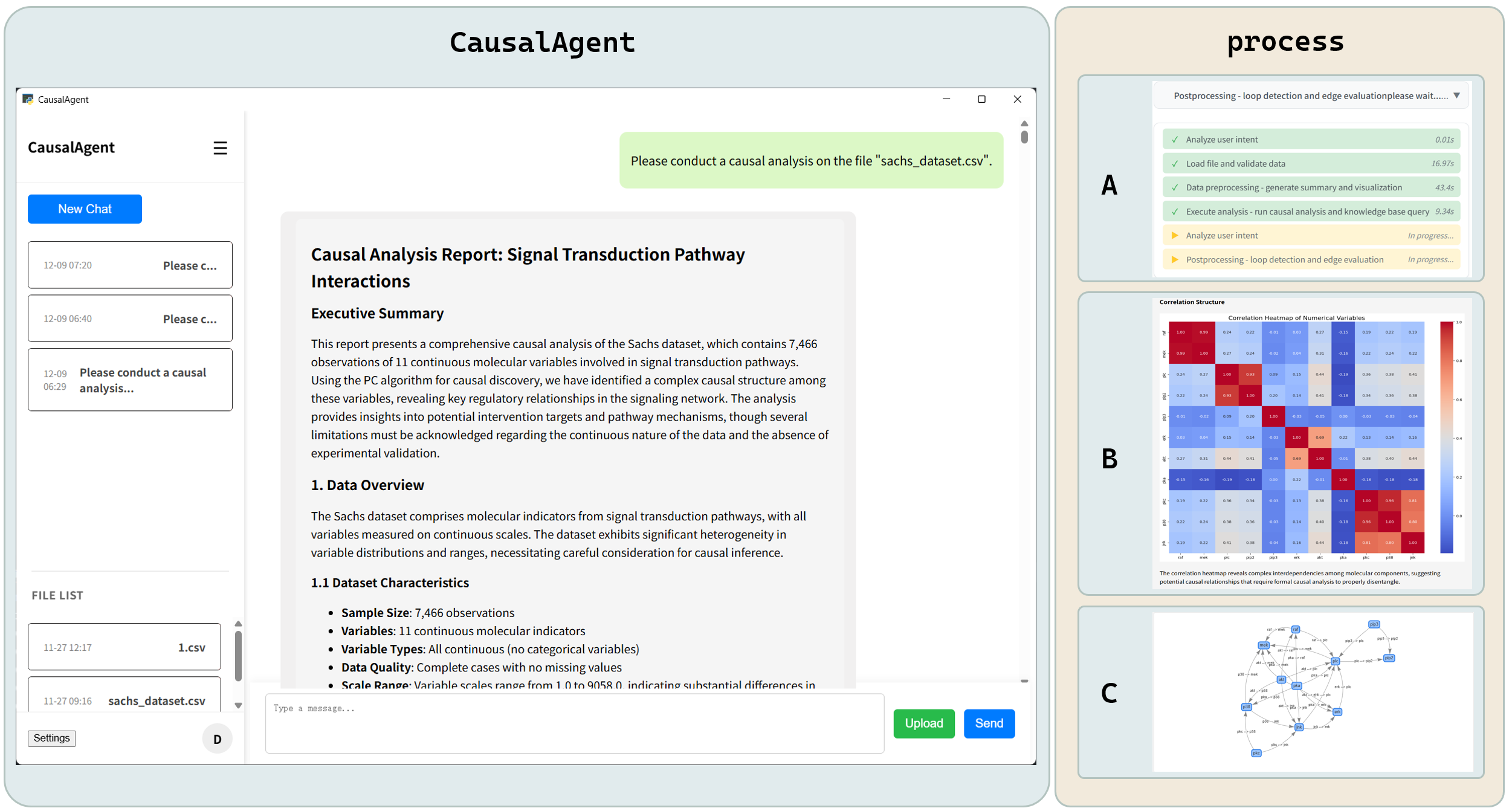} 
  \caption{Demonstration of CausalAgent on the Sachs Dataset. (A) CausalAgent Running process. (B) Correlation heatmap of variables. (C) Generated Causal diagram.}
  \Description{The figure is divided into two main sections illustrating the CausalAgent system. 
The left section shows the main user interface, featuring a chat window on the left sidebar and a file list at the bottom. In the main content area, a user prompt ``Please conduct a causal analysis on the file `sachs\_dataset.csv'" is visible, followed by a generated ``Causal Analysis Report" that includes an Executive Summary and Data Overview. 
The right section, labeled "process," breaks down the analysis into three vertically arranged visualization steps labeled A, B, and C. 
Panel A shows a progress log with green check marks indicating completed tasks such as ``Load file and validate data" and "Data preprocessing." 
Panel B displays a correlation heatmap with a red-and-blue grid representing the statistical relationships between numerical variables. 
Panel C presents the final output: a directed acyclic graph (DAG) showing nodes connected by arrows, representing the learned causal structure.}
  \label{fig:fig2}
\end{figure*}

\subsection{Reporting Agent}
    By integrating RAG \cite{lewis2020rag}, this agent synthesizes diagnostic metrics and causal structures into comprehensive reports. These reports comprehensively cover methodological details, key causal findings, and model limitations. Furthermore, the system exports the graph to an interactive frontend component, allowing users to explore complex dependencies through visual operations like node manipulation and zooming.

\section{Key Implementation}
To ensure robustness, we employ the following four strategies:
\begin{enumerate}[(1)]
    \item \textbf{Retrieval-Augmented Generation (RAG).} We constructed a vectorized knowledge base from causal textbooks to ground LLM explanations in established theory.
    \item \textbf{Supervised Fine-Tuning (SFT).} The base model was fine-tuned to enhance instruction-following and accuracy in translating statistical results into natural language.
    \item \textbf{Prompt Engineering \& MCP.} Adopting MCP \cite{anthropic2024mcp} standardizes the interface between LLMs and external tools, decoupling reasoning from execution.
    \item \textbf{LLM Backbone.} We utilize GLM-4.6 \cite{glm4.6}, leveraging its reasoning capabilities for task distribution and conflict resolution.
\end{enumerate}

\section{Demonstration Scenario}

To demonstrate the efficacy and interactivity of CausalAgent in scientific discovery, we present a case study using the widely-cited Sachs Protein Signaling Dataset \cite{sachs2005causal}. This dataset consists of flow cytometry measurements of 11 phosphorylated proteins and phospholipids (e.g., Raf, Mek, Erk) derived from human immune system cells, serving as a gold standard for validating causal structure learning algorithms.

\subsection{Workflow Walkthrough}
As shown in Figure \ref{fig:fig2}, the analysis workflow proceeds as follows:

\textbf{Step 1: Data Upload and Query.}
The user uploads the \textit{sachs.csv} file and inputs a high-level directive: \textit{``Please conduct a causal analysis on the file `sachs\_dataset.csv'."}

\textbf{Step 2: Multi-Agent Collaboration.}
CausalAgent automatically profiles the dataset. By analyzing data distribution characteristics within this biological context (dense network), it autonomously selects the PC algorithm for pathway reconstruction—without requiring the user to interpret complex raw adjacency matrices.

\textbf{Step 3: Report Generation.}
Following Processes B and C depicted in Figure 2, CausalAgent synthesizes algorithmic outputs with its built-in knowledge base to generate a comprehensive analysis report. The report identifies \textit{Akt} and \textit{Pka} as master regulators, explicitly recommending them as priority targets for drug intervention studies.

\textbf{Step 4: Interactive Refinement.}
Following the report, the user asks a follow-up question: \textit{``What if we intervene on Mek? How would Erk change?"}. The system responds by synthesizing the \textbf{derived causal analysis results} with \textbf{domain priors}.

\section{Conclusion and Future Work}
We presented CausalAgent, a conversational multi-agent system for end-to-end causal inference. The system integrates MAS, RAG, and multiple causal algorithms encapsulated by the MCP protocol. While ensuring high extensibility, it provides users with a convenient interactive experience, significantly reducing the barrier to the causal domain and providing a better platform for expanding causal inference development.

Future work will focus on two directions: first, at the algorithmic level, integrating structure learning methods supporting Latent Confounders and counterfactual reasoning modules to achieve quantitative causal effect estimation; second, regarding domain adaptation, introducing expert feedback mechanisms for high-risk scenarios like healthcare and finance to further enhance system safety and reliability.

\begin{acks}
This research was supported in part by the National Science and Technology Major Project (2021ZD0111502), the Natural Science Foundation of China (U24A20233, 62206064), the National Science Fund for Excellent Young Scholars (62122022), the Guangdong Basic and Applied Basic Research Foundation (2025A1515010172), and the Guangzhou Basic and Applied Basic Research Foundation (2024A04J4384).
\end{acks}

\section*{GenAI Usage Disclosure}
Generative AI tools (Gemini, Gemini-3.0-pro) were used to improve the readability of the manuscript and assist in prototyping the agent interaction logic. The entire content, including code and text, was manually verified by the authors to ensure scientific rigor.
\bibliographystyle{ACM-Reference-Format}
\bibliography{references}

\end{document}